\documentclass[letterpaper, 10 pt, conference]{ieeeconf}  

\IEEEoverridecommandlockouts                              
\usepackage[linesnumbered,ruled,vlined]{algorithm2e}                                                   
\usepackage{amsmath,amssymb,amsfonts}
\usepackage{algorithmic}
\usepackage{lmodern}
\usepackage{mathtools}
\usepackage{url}

\usepackage{breakurl}
\usepackage[breaklinks]{hyperref}
\usepackage{multirow}
\usepackage{booktabs}
\usepackage[]{footmisc}   
\usepackage{tabularray}
\usepackage{array}
\usepackage{makecell}
 \usepackage{graphicx}
 \usepackage{multirow}
 \usepackage{booktabs}

\title{\LARGE \bf
Trajectory Prediction with Observations of Variable-Length for Motion Planning in Highway Merging Scenarios*
}

\author{Sajjad Mozaffari, MReza Alipour Sormoli, Konstantinos Koufos, Graham Lee, and Mehrdad Dianati$^{2}$
\thanks{*This work was part of the Hi-Drive project. The project has received funding from the European Union's Horizon 2020 research and innovation programme under grant agreement No. 101006664.}
\thanks{
1. 
2. All authors are with WMG,
        The University of Warwick, Coventry, U.K.
        {\tt\small Email: \href{mailto:sajjad.mozaffari@warwick.ac.uk}{sajjad.mozaffari@warwick.ac.uk} }}%
}

\begin{document}

\maketitle
\thispagestyle{plain}
\pagestyle{plain}

\begin{abstract} 
Accurate trajectory prediction of nearby vehicles is crucial for the safe motion planning of automated vehicles in dynamic driving scenarios such as highway merging. Existing methods cannot initiate prediction for a vehicle unless observed for a fixed duration of two or more seconds. This prevents a fast reaction by the ego vehicle to vehicles that enter its perception range, thus creating safety concerns. Therefore, this paper proposes a novel transformer-based trajectory prediction approach, specifically trained to handle any observation length larger than one frame. We perform a comprehensive evaluation of the proposed method using two large-scale highway trajectory datasets, namely the highD and exiD. In addition, we study the impact of the proposed prediction approach on motion planning and control tasks using extensive merging scenarios from the exiD dataset. To the best of our knowledge, this marks the first instance where such a large-scale highway merging dataset has been employed for this purpose.  The results demonstrate that the prediction model achieves state-of-the-art performance on highD dataset and maintains lower prediction error w.r.t. the constant velocity across all observation lengths in exiD. Moreover, it significantly enhances safety, comfort, and efficiency in dense traffic scenarios, as compared to the constant velocity model.
\end{abstract}

\section{Introduction} 
The successful development of highly automated driving systems in interactive driving scenarios requires understanding the potential evolution of the driving situation in the immediate future. A prime example of such a scenario is merging onto highways, particularly during peak hours. In this context, autonomous vehicles (AVs) must continually adjust their motion based on the trajectory predictions of other road users on the main carriageway. Recent progresses in deep learning have facilitated the development of advanced interaction-aware prediction models that exhibit significantly lower prediction errors and extended prediction horizons~\cite{mozaffari2020deep}. However, there remains a lack of research to investigate the practical limitations and the impact of these methods on downstream motion planning and control in automated vehicles.

One drawback of existing learning-based prediction approaches is their reliance on observing the states of nearby vehicles for a fixed duration, typically two seconds or more~\cite{sheng2022graph, schmidt2022crat, deo2018convolutional}. That delays the availability of predictions for the first few seconds of the first time when a vehicle enters the field of view of the AV's perception system. These delays in prediction can hinder timely and safe decision-making by AVs, especially in highly interactive driving scenarios.

In addition, many existing trajectory prediction approaches lack a thorough analysis of how the learning-based prediction impacts downstream motion planning and control tasks. Even though some studies~\cite{chen2022scept, wang2021risk, tang2022prediction} offer qualitative analysis of motion planning using prediction data, they often fall short of providing a comprehensive comparison between learning-based predictions and conventional methods like the constant velocity prediction model. Additionally, these evaluation methods often neglect to adequately assess the influence of prediction quality in diverse driving conditions, such as high/low traffic densities.  To the best of our knowledge, there is a notable absence of studies evaluating the impact of learning-based prediction on motion planning and control for merging onto highways.

To address the aforementioned shortcomings, this paper first proposes a new formulation of vehicle trajectory prediction where a variable-length sequence of past observations is employed as opposed to the fixed duration used in existing methods. Then, we design a novel transformer-based prediction model using a tailored input feature list for highway merging scenarios. The proposed model is specifically trained to handle variable-length observations.  The prediction data is then utilised in motion planning and control for vehicles merging onto highways. To this end, the prediction data is encoded into a potential field and fed to a Model Predictive Control (MPC)-based motion planning and control algorithm. We evaluate the performance of the prediction model and the prediction-based motion planning using a large-scale highway merging dataset, namely the exiD dataset~\cite{exiDdataset}.
Additionally, we separately evaluate the performance of the prediction model using a benchmark highway driving dataset, namely the highD dataset~\cite{krajewski2018highd}, to facilitate comparison with several state-of-the-art highway trajectory prediction models. The contributions of this paper can be summarised as follows:
\begin{itemize}
    \item A novel formulation for vehicle trajectory prediction that accommodates variable-length observations.
    \item A novel transformer-based vehicle trajectory prediction model incorporating interaction and map-aware features for highway merging scenarios.
    \item Statistical evaluation of the prediction model and its impact on downstream motion planning and control in highway merging scenarios.
\end{itemize}


\section{Related Work}~\label{sec:rel_work} 
This section reviews the existing studies on trajectory prediction, particularly focusing on two key aspects: (1) Various learning-based techniques employed for trajectory prediction, and (2) studies that analyse the impact of nearby vehicle trajectory prediction on the motion planning of the Ego vehicle (EV).

\subsection{Learning-based Trajectory Prediction}
Recent learning-based methods has been reviewed based on their input representation, prediction model and output type in~\cite{mozaffari2020deep}. Recurrent Neural Networks (RNNs), particularly Long Short-Term Memories (LSTMs), have been widely utilized for vehicle trajectory prediction~\cite{altche2017lstm, deo2018multi, park2018sequence, messaoud2020attention}. However, with the introduction of the attention mechanism~\cite{vaswani2017attention}, RNNs are gradually being outperformed in many sequence-to-sequence tasks, such as speech recognition~\cite{gulati2020conformer} and natural language processing~\cite{vaswani2017attention}. Additionally, Convolutional Neural Networks (CNNs) are employed in vehicle trajectory prediction due to their advantages in spatial interaction learning~\cite{deo2018convolutional, cui2019multimodal}. 

More recently, transformer neural networks have emerged as an alternative to both RNNs and CNNs~\cite{giuliari2021transformer, huang2022multi, gao2023dual}. Giuliari et al.~\cite{giuliari2021transformer} demonstrated that simple transformer neural networks outperform state-of-the-art LSTM prediction models in human trajectory prediction tasks. In~\cite{huang2022multi}, the multimodal vehicle trajectory prediction task was addressed using transformers where each predicted trajectory is conditioned on a separate attention head. Gao et al.~\cite{gao2023dual} employed dual transformers to predict both the intention and the trajectory of a target vehicle leveraging vehicle interaction features and the history of the lateral states of the target vehicle. In~\cite{Mozaffari2022}, a transformer encoder was used to learn latent representation from a list of interaction-aware, context-aware, and dynamic-aware input data. That representation was used to predict multimodal manoeuvres and trajectories for a target vehicle. Similarly, in this paper, we employ a transformer neural network using input features specially designed for merging scenarios. Notably, unlike existing studies that require a fixed observation length of two to three seconds, our proposed method can predict a vehicle's trajectory with a minimum of two time-step observations (equivalent to 0.4 seconds).

\subsection{Trajectory Prediction for Motion Planning}

Wang et al.~\cite{wang2021risk} designed an MPC-based motion planner for risk mitigation leveraging trajectory prediction using LSTMs. The authors investigated the impact of constant velocity prediction, non-linear input-output neural network and LSTM on the performance of motion planning. The LSTM model is trained and evaluated on the highD~\cite{krajewski2018highd} dataset, however, the planning algorithm is evaluated using two selected scenarios from the highD dataset. Tang et al.~\cite{tang2022prediction} expanded upon this work by estimating the uncertainty of prediction using a deep ensemble technique. Then, the effects of uncertainty-aware prediction on motion planning are evaluated in a single cut-in and merging scenario. Building upon these studies, we further extend the research field by conducting a comprehensive statistical analysis of the planning algorithm's performance.

In~\cite{chen2022scept, cui2021lookout}, multimodal joint prediction of nearby vehicles' trajectories is used for contingency planning. Chen et al.~\cite{chen2022scept} utilised an MPC to plan a trajectory for each prediction mode, and then the results are discussed for a specific driving scenario qualitatively. Cui et al.~\cite{cui2021lookout} used a sampling-based planning approach in simulation with reactive agents. The simulation scenarios are initialised with real-world urban data. In~\cite{jeong2020surround}, an LSTM neural network is used to predict the trajectory of the EV's nearby vehicles in a multi-lane turn intersection, and then the future motion of the EV is planned using the prediction data and MPC planner. The results of the planning algorithm are compared with a path following and constant turning rate and velocity predictions. Different from these studies, we target highway merging scenarios and analyse the impact of trajectory prediction of other vehicles on joint motion planning and control of merging vehicles. The impact is studied for learning-based and constant velocity-based predictors and for various levels of traffic.
\begin{figure*}[!t] 
\centering
\includegraphics[width=6.8in]{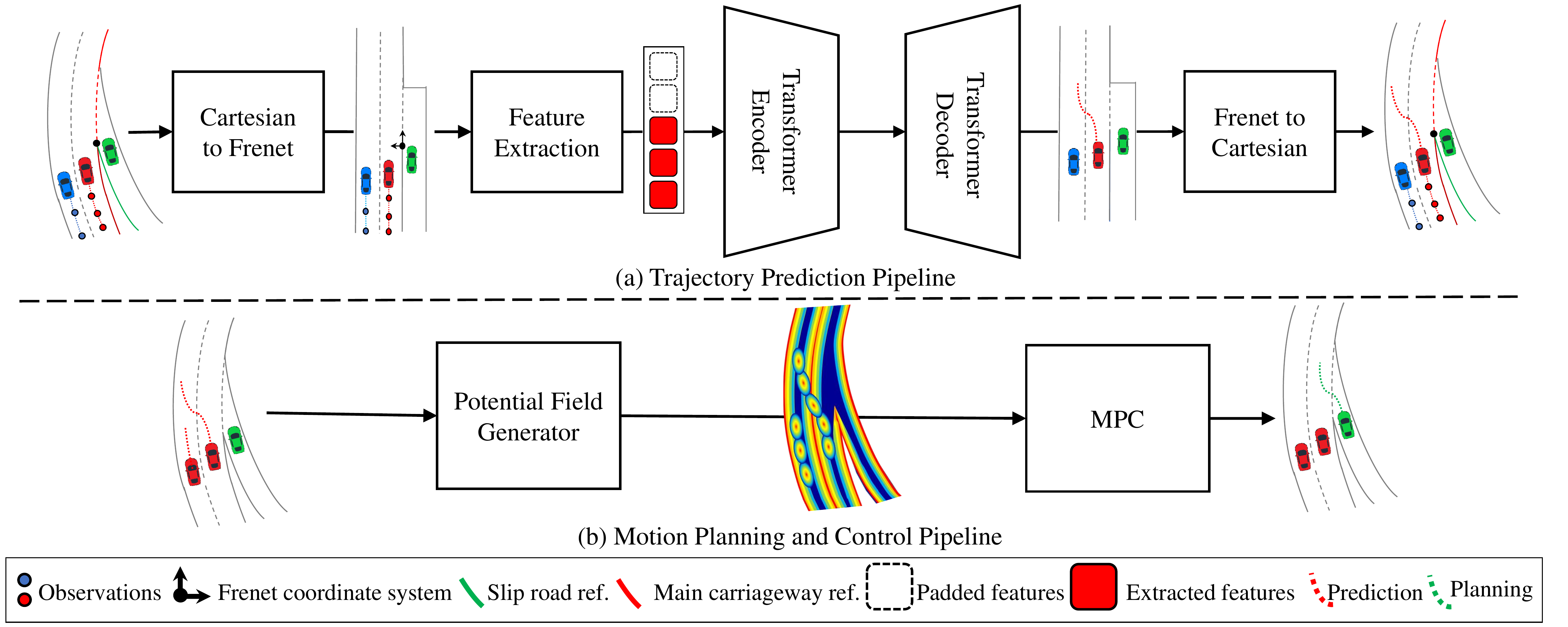}
\caption{An overview of the proposed prediction and motion planning pipelines with an illustrative exemplar scenario. Note that the example output of the Potential Field Generator box is cumulatively drawn over several prediction steps for illustrative purposes. }
\label{fig:model}
\end{figure*}

\section{Methodology}\label{sec:method}
This section first defines the vehicle trajectory prediction problem and the system model. Then describes the proposed trajectory Prediction model with Observation of Variable Length (POVL). Finally, it discusses the motion planning and control algorithm. Fig.~\ref{fig:model} illustrates an overview of the prediction and planning pipelines using an exemplary driving scenario.

\subsection{System Model and Problem Definition}\label{sec:prob_def} 
We consider a semi- or fully-automated Ego Vehicle (EV) merging from a single-lane slip road into a main carriageway with an arbitrary number of lanes. All roadways can be straight or curvy. The EV intends to predict the future trajectory of vehicles along both the main carriageway and the slip road, and these predictions are subsequently fed into its motion planning and control module. The tracking data of these vehicles and the lane markings are assumed to be available for the EV, e.g., through onboard perception, V2X communication, or cloud. Assuming that the future trajectories of vehicles are independent, the EV predicts the trajectory of one Target Vehicle (TV) at a time and for all perceived vehicles in the scene.

The problem of trajectory prediction is defined as estimating the sequence of TV's future x-y positions during $T_{pred}$ time-steps of prediction window $Y_{TV} =\{(x_1,y_1),\ldots (x_{T_{pred}}, y_{T_{pred}})\}$ given the states of the TV and its surrounding environment during $T_{obs}$ time-steps of the observation window. The observation length for a TV can vary between a minimum and maximum value, i.e., $T_{obs}\in [T_{min}, T_{max}]$.

\subsection{Prediction model} 
Fig.~\ref{fig:model}-a illustrates the overview of the prediction framework. The processing steps are summarised below:
\begin{enumerate}
\item The TV's tracking data is converted to the Frenet coordinate system.
\item The input features are computed using the tracking and map data.
\item Transformer encoder-decoder neural network is used to predict the future trajectory of the TV.
\item The prediction data are converted back to the Cartesian coordinate system.
\end{enumerate}
\subsubsection{Coordinate Conversion}
The prediction is conducted in the Frenet coordinate system, which describes the movement of vehicles in terms of along-track ($\vec{s}$) and cross-track ($\vec{d}$) components. As the curvature of the slip road may differ from that of the main carriageway, two distinct reference paths are considered for vehicles on each road segment, as illustrated in Fig.~\ref{fig:model}. The crossing point of the reference paths serves as the origin of the new coordinate system. A vehicle is first associated either with the slip road or the main carriageway reference path\footnote{Note that it is assumed vehicles do not deviate from the road boundaries.}. Afterwards, its cross-track component is determined by calculating the distance of the orthogonal projection between the vehicle's centre and the associated reference path. The along-track component is calculated as the arc length along the reference path from the origin to the orthogonal projection point of the vehicle's centre. Similarly, the prediction data are converted back from Frenet to Cartesian coordinate system.
\begin{figure}[!t] 
\centering
\includegraphics[width=3.4in]{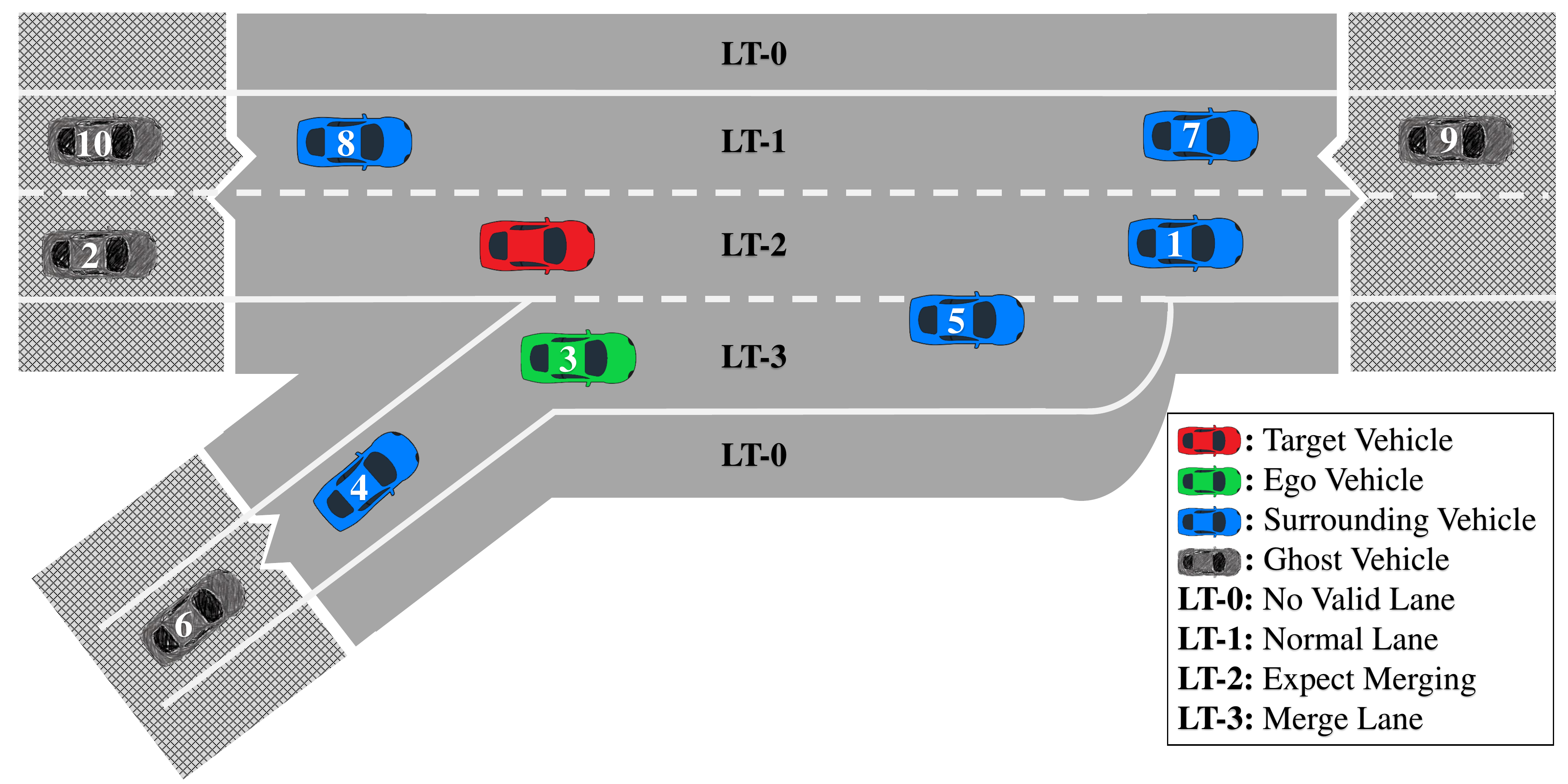}
\caption{Illustration of the maximum ten surrounding vehicles for a target vehicle and different lane types (best seen in colour).}
\label{fig:features}
\end{figure}

\subsubsection{Input Features}
The input to the prediction model consists of a sequence of feature lists with a maximum observation length of $T_{max}$ time steps. If there are less than $T_{max}$ observations available for a vehicle, the rest of the sequence is padded with zero values. 

The observation at each time step is represented by a list of 28 features, categorized into three groups. The first group describes the motion of the TV, encompassing features including the TV's lateral position in the lane, longitudinal velocity, and lateral and longitudinal acceleration. The second group characterizes the TV's interaction with surrounding vehicles (SVs), considering the lateral and longitudinal distance between the TV and each SV. A maximum of 10 SVs are considered in this study including the 
preceding/following vehicles on the TV's lane (1/2), the right close preceding/following vehicles (3/4), right far preceding/following vehicles (5/6), left close preceding/following vehicles(7/8), and left far preceding/following vehicles(9/10). A ghost vehicle at far distances replaces the feature list of each SV that doesn't exist or is not in the perception range of the EV. Fig.~\ref{fig:features} shows the numbered labelling of SVs for an example merging scenario. Finally, the third group of features pertains to the driving environment, including lane width and the types of right and left lanes relative to the TV. This study defines four lane types: (0) no lane, (1) normal lanes (where crossings are anticipated into and from them), (2) expect merging lanes (where crossings are expected into them), and (3) merge lane (where crossings are expected from them). Fig.~\ref{fig:features} provides an illustration of the surrounding vehicles and the various lane types.
\begin{figure}[b!]
\centering
\includegraphics[width=.5\linewidth]{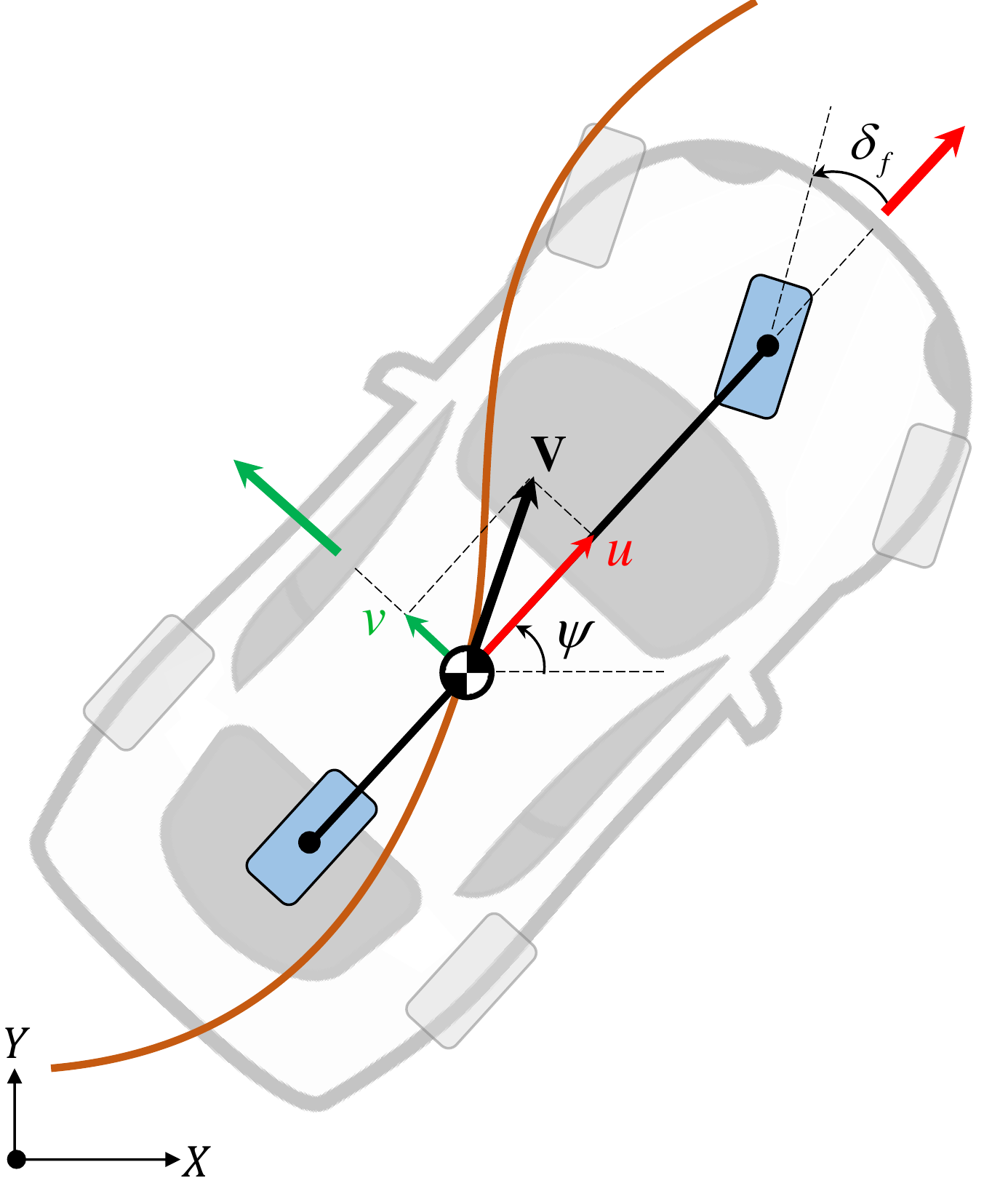}
\caption{Kinematic variables and parameters of the bicycle model used in the equation of motion.}
\label{fig: dynamic model}
\centering
\end{figure}

\subsubsection{Transformer Model}
The prediction model includes a transformer encoder and decoder components, initially introduced in~\cite{vaswani2017attention}. Transformer neural networks utilise multi-head self-attention and cross-attention mechanisms to identify and focus on the informative segment of the input sequence in alignment with the underlying training objectives. Interested readers are referred to~\cite{vaswani2017attention} for further details about transformers.

The transformer encoder is used to encode the variable length observation into a sequence of latent representations capturing the spatiotemporal dependencies of the input sequence. The transformer decoder then utilises the latent representations to predict the displacement of the TV at each time step during the prediction window $T_{pred}$. The masking mechanism of transformer encoders and decoders is used to filter out the padded sequence of input data.

The encoder and decoder models consist of two transformer layers each with eight attention heads, a model dimension of 512, and a feed-forward hidden layer dimension of 128. To represent the uncertainties in the prediction output, the model estimates the parameters of a bivariate Gaussian distribution of displacement in along-track and cross-track dimensions for each prediction step.

\subsection{Motion Planning Algorithm} 
In this section, we design an optimization-based motion planning method for the EV attempting to merge into the highway based on the receding horizon algorithm a.k.a MPC. The algorithm receives the predicted future trajectories of the dynamic objects in the driving environment and calculates the optimized trajectory for the EV by minimizing a given cost function~\cite{rasekhipour2016potential}. Specifically, the optimization cost at a time step $t$ is a function of the EV  dynamics/constraints ($U^{ev}$), the perceived environment represented by a potential field ($U^{env}$), and a reference signal ($U^{ref}$) over the planning horizon $p$. One may write 
\begin{equation}
\label{eq: cost}
    Cos{t_t} = \sum\limits_{i = 0}^p {U_{t + i}^{ev} + U_{t + i}^{env} + U_{t + i}^{ref}}.
\end{equation}
Each term of the cost function is further explained in the subsequent sections.
\subsubsection{Vehicle Dynamics}
In this paper, the (dynamic) bicycle model has been adopted for modelling the dynamic motion of the EV. 
According to it, the EV motion is governed by a set of nonlinear equations~\cite{rasekhipour2016potential}: 
 
\begin{equation}
    \label{eq: nonlinear_short}
    \begin{array}{l}
\dot x = f\left( {x,{u_c}} \right)\\
x = {\left[ {\begin{array}{*{20}{c}}
X&u&Y&v&\psi &{\dot \psi }
\end{array}} \right]^T}\\
{u_c} = {\left[ {\begin{array}{*{20}{c}}
{{F_u}}&\delta_f 
\end{array}} \right]^T},
\end{array}
\end{equation}
where $x$ and $u_c$ are the state vector and control effort (inputs) respectively, and $f\left( \cdot  \right)$ represents a set of six nonlinear functions. With reference to the coordinate system in Fig.~\ref{fig: dynamic model}, the state vector $x$, includes the parameters $u$ and $v$ for the longitudinal and lateral velocities of the EV, $\psi$ and $\dot \psi$ are the heading and yaw rate, and the Cartesian coordinates $(X,Y)$ represent the location (centre of mass) of the EV. Moreover, the input vector $u_c$ consists of throttle/brake force ($F_u$) and the front wheel steering angle ($\delta_f$).

Next, the state equations are adaptively linearised around their operating point to facilitate the optimisation process yielding the following simplified form of state equations: $\dot x = Ax + B{u_c}$,
where $A$ and $B$ are the Jacobians of $f\left( {x,{u_c}} \right)$ with respect to $x$ and $u_c$, respectively. Finally, the term $U^{ev}$ in Eq.~\ref{eq: cost} is defined as the quadratic weighted sum of the control signals and their rate of change 
\begin{equation}
\label{eq: u_ev}
    {U^{ev}} = u_c^T{Q_1}{u_c} + \dot u_c^T{Q_2}\dot {u_c}, 
\end{equation}
where $Q_1$ and $Q_2$ are constant square matrices (zero non-diagonal components) to tune the contribution of the control signal and its rate in the overall cost.
\subsubsection{Potential field representation of driving environment}
Repulsive potential fields have been widely adopted in path planning of autonomous vehicles (AVs) for modelling the interactions between the EV and other elements in the driving scene such as obstacles and boundaries. A repulsive potential field guides the EV away from the associated obstacle/boundary; by carefully designing the potential function it becomes possible to differentiate between different types of obstacles, which can be critical from a safety point of view. For instance, in a highway merging scenario, the main obstacles that should be avoided can be broadly categorized into three types: Moving/static objects such as other vehicles, road boundaries (non-crossable obstacles), and lane markings (crossable obstacles). Next, a repulsive potential is designed for each category to build the overall potential field representing the driving context. Intuitively, the maximum of a potential field should be taken at the obstacle's location.

The potential field for each vehicle, $V_o$, can be defined based on its distance to the EV after this is appropriately weighted in the lateral and longitudinal directions.
\begin{equation}
\label{eq:Vo}
    V_o = \frac{{{a_o}}}{{{{\left( {\frac{{x - {x_{obs}}}}{X_c}} \right)}^{{c_x}}} + {{\left( {\frac{{y - {y_{obs}}}}{Y_c}} \right)}^{{c_y}}}}}, 
\end{equation}
where $(x_{obs},y_{obs})$ are the obstacle location, $a_o$ is the obstacle vehicle potential field magnitude, $X_c, Y_c$ are constants determining the rate of change for the potential field in x and y directions, and $c_x, c_y$ is another set of weights allowing to model every vehicle in the 2d plane by an ellipse instead of a circle. One can find in Fig.~4~(top), an example illustration.  

The potential field associated with road boundaries, $V_r$, is defined according to the EV's distance to the boundary, $d_r$. In that case, the potential should be high when the EV approaches near the boundary, but can be degenerated to zero when the distance $d_r$ becomes larger than a threshold, $D_r$. Specifically, 
\begin{equation}
    V_r = \left\{ {\begin{array}{*{20}{c}}
{{a_r}{{\left( {{d_r} - {D_r}} \right)}^2}, \quad {d_r} \leq {D_r}}\\
{0, \quad \quad \quad \quad \quad \;\;{d_r} > {D_r}}, 
\end{array}} \right.
\end{equation}
with $a_r>0$ being the maximum of the potential field $V_r$. 

 Finally, the lane-marking potential, $V_l$, may have a concave shape with its magnitude $a_l$ being lower than the other two categories, i.e., $a_l<a_o$ and $a_l<a_r$ as can also be seen in Fig.~4, since lane crossing should be permitted for overtaking.  
\begin{equation}
V_l = {a_l}\exp \left( { - {b_l}d_l^2} \right), 
\label{eq:Vl}
\end{equation}
where $d_l$ is the distance to the lane marks and $b_l>0$ determines the rate of change. 

 Finally,  the environment cost for each time step in Eq.~\ref{eq: cost} is obtained by adding up the terms calculated in Eq.~\eqref{eq:Vo} - \eqref{eq:Vl}:
\begin{equation}
    U^{env} = V_r + V_l + \sum\nolimits_o V_o, 
\end{equation}
where the summation is over all obstacle vehicles.

\begin{figure}[!t] 
\centering
\includegraphics[width=3.4in]{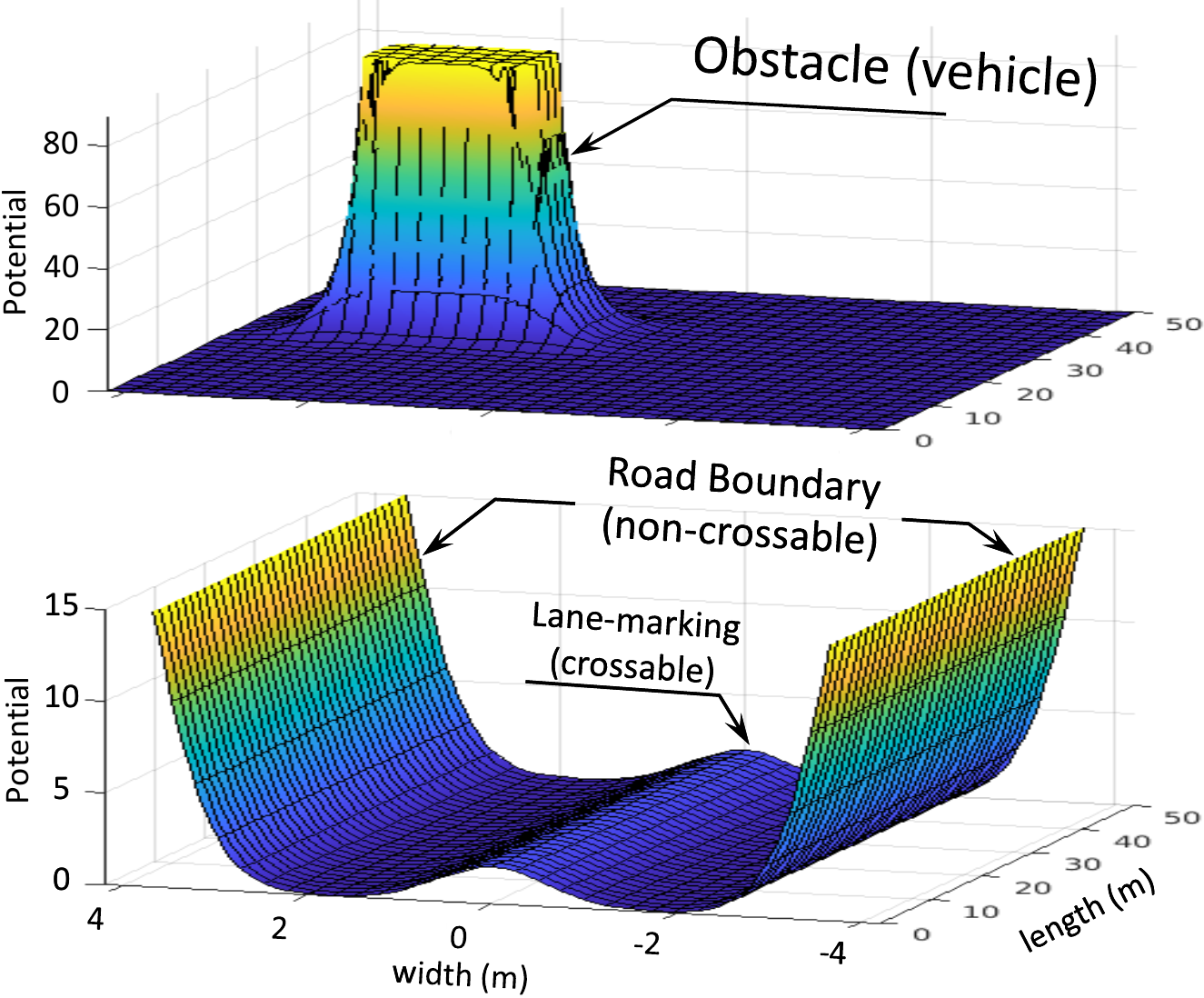}
\caption{Potential field for modelling semantic information in the driving environment such as other vehicles, road boundaries and lane markings (the axes have been scaled).}
\label{fig: pf}
\end{figure}

\subsubsection{Reference signal}
The last term of the cost function (Eq.~\ref{eq: cost}) is calculated based on two reference signals, namely the distance from the center-line of the desired merging lane ($y_{des}$) and the nominal speed of that lane ($u_{des}$). This term is essential to guarantee progress in the EV's trajectory, otherwise, it is possible that the EV stops moving to minimize the rest of the terms in the cost function. Accordingly, the contribution of the reference signals in the overall cost value is: 
\begin{equation}
    \label{eq: ref signal}
    {U^{ref}} = {Y^T}{Q_3}Y, \quad \quad \;Y = {\left[ {\begin{array}{*{20}{c}}
{{y_{des}}}&{{u_{des}}}
\end{array}} \right]^T},
\end{equation}
where $Q_3$ is a constant square matrix (zero non-diagonal components) to tune the contribution of the reference signal error in the final cost. 

In the end, the optimal control action at time step $t$ is obtained by minimizing the cost function including $p$ future time steps as in Eq.~\ref{eq: cost}. Sequential quadratic programming is used to solve to the optimum input over the control horizon.  

\section{Performance Evaluation}~\label{sec:eval}
This section offers a performance evaluation of the proposed prediction model and analyzes its impact on motion planning and control algorithms. To this end, it first presents the datasets used for evaluation, followed by the implementation details. Then, we provide the selected prediction and planning metrics. Subsequently, the results of prediction and planning models are discussed.

\subsection{Dataset} 
\subsubsection{exiD Dataset} We utilise the exiD dataset~\cite{exiDdataset} for both trajectory prediction and motion planning evaluation in highway merging scenarios. This dataset comprises a large-scale collection of naturalistic vehicle trajectories at highway entries and exits, recorded by drones in Germany in 2022. Among seven different locations reported in the dataset, the merging data of four locations is used in this study. These locations feature a single-lane slip road merging into two- or three-lane main roads. The selected data contains 49 recordings from which 45 are used for training and 4 for evaluation of the prediction model, one per location. Recording number 39, from the prediction test set, is used for evaluating the planning algorithm.

\subsubsection{highD Dataset:} The prediction model is also separately trained and evaluated on the highD dataset~\cite{krajewski2018highd}, which is a benchmark highway driving dataset used in several trajectory prediction studies~\cite{messaoud2020attention, gao2023dual, mozaffari2020deep}. The highD dataset contains 110,000 vehicle trajectories from six different highways in Germany. The dataset is recorded using a drone from mainly straight highways and includes different levels of traffic densities. The dataset is divided into the train, validation, and test sets with the ratio of $70\%$, $10\%$, and $20\%$, following the experimental protocol of existing studies~\cite{Tang2019}. Both highD and exiD datasets are reported with 25 frames per second which are reduced to 5 in the prediction task.

\subsection{Implementation Details}
The prediction model is trained using Adam optimiser~\cite{kingma2014adam} with a learning rate of $0.0001$ for a maximum of $10^5$ batches, each with a size of 2000 samples. Observations ranging from a minimum of two (0.4 seconds) to a maximum of 15 timesteps (3 seconds) are used to predict the next 25 timesteps (5 seconds). The prediction model is implemented using PyTorch library~\cite{paszke2017automatic} and is run on GeForce RTX 2080 TI GPU. The source code of this study is available at \href{https://github.com/SajjadMzf/TrajPred}{https://github.com/SajjadMzf/TrajPred}

The planning algorithm is evaluated for large-scale highway merging scenarios initiated from realistic data from the exiD dataset. Each scenario has a five-second planning horizon and is initialised using different time steps of a merging vehicle. During each time step of the scenario, the motion planning algorithm uses the corresponding time step of prediction of other vehicles. In total 97 merging vehicles are used for motion planning evaluation. The planning algorithm is implemented in MATLAB.

\subsection{Evaluation Metrics} 
\subsubsection{Prediction Metric}  The Root Mean Square Error (RMSE) between the predicted trajectory of vehicles and their ground truth is used to evaluate the accuracy of the prediction model: 
\begin{equation}
    \label{eq: metrics}
    \begin{array}{l}
{\textnormal{RMSE}} = \sqrt {\frac{1}{N}\sum\limits_{i = 1}^N {\left[ {{{\left( {z_i - {z_{i,gt}}} \right)}^2} + {{\left( {w_i - {w_{i,gt}}} \right)}^2}} \right]} }, 
\end{array}
\end{equation}
where $(z_i,w_i)$ are the predicted (Cartesian) coordinates at the $i$-th sample, out of N total samples, and $(z_{i,gt},w_{i,gt})$ is the associated ground truth.

\subsubsection{Planning Metrics}
The motion planning algorithm is evaluated based on three criteria, namely, safety, comfort, and efficiency. The inverse of the average Time-To-Collision, $iTTC$, is reported to evaluate the safety of the planned trajectories, the average Jerk represents the comfort, and the average force  is reported for fuel efficiency.

\begin{table}
\centering
\caption{Comparision of baseline studies using RMSE(m) at different prediction horizons evaluated on highD and exiD Dataset. The Best and second best errors are marked as \textbf{bold} and \underline{underlined}.}
\label{tab:comp_highd}
\resizebox{\linewidth}{!}{%
\begin{tblr}{
  cell{2}{1} = {r=5}{},
  cell{7}{1} = {r=2}{},
  hline{1,9} = {-}{0.08em},
  hline{2,7} = {-}{},
}
                          Dataset          & Model       & 1s   & 2s   & 3s   & 4s   & 5s   \\
highD                               & CV & \underline{0.10} & 0.33 & 0.69 & 1.17 & 1.76 \\
                                    & CS-LSTM~\cite{deo2018convolutional}    & 0.19 & 0.57 & 1.16 & 1.96 & 2.96 \\
                                    & MHA-LSTM~\cite{messaoud2020attention}   & \textbf{0.06} & \textbf{0.09} & \underline{0.24} & \underline{0.59} & \underline{1.18} \\
                                    & Dual Trans.~\cite{gao2023dual} & 0.41 & 0.79 & 1.11 & 1.40 & -    \\
                                    & POVL (ours) & 0.12 & \underline{0.18} & \textbf{0.22} & \textbf{0.53} & \textbf{1.15} \\
exiD   & CV  & 0.25 & 0.63 & 1.19 & 1.92 & 2.82 \\
                                    & POVL (ours) & \textbf{0.17} & \textbf{0.41} & \textbf{0.76} & \textbf{1.21} & \textbf{1.75} 
\end{tblr}
} 

\end{table}

\begin{figure}[!t] 
\centering
\includegraphics[width=3.4in]{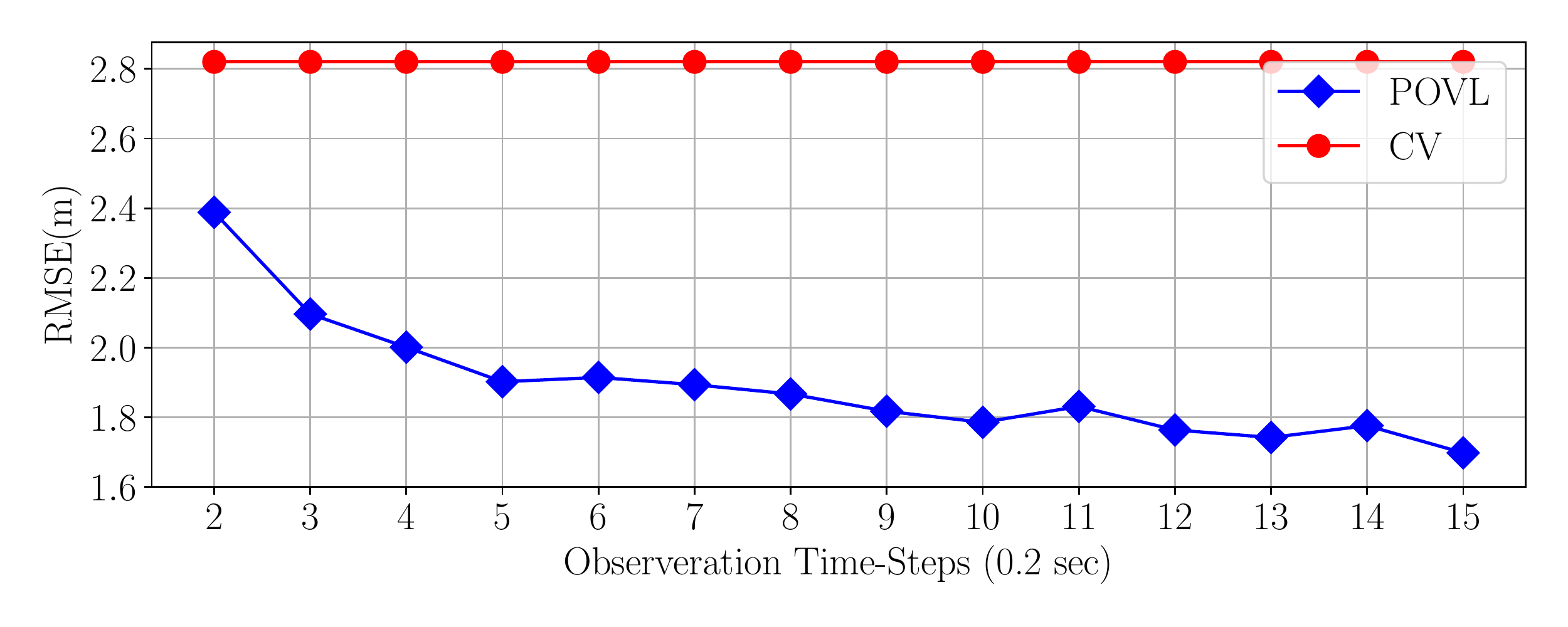}
\caption{RMSE of the proposed prediction model in different observation lengths on exiD dataset.}
\label{fig:ovl}
\end{figure}

\subsection{Prediction Results} 
\begin{figure*}[!t] 
\centering
\includegraphics[width=.9\linewidth]{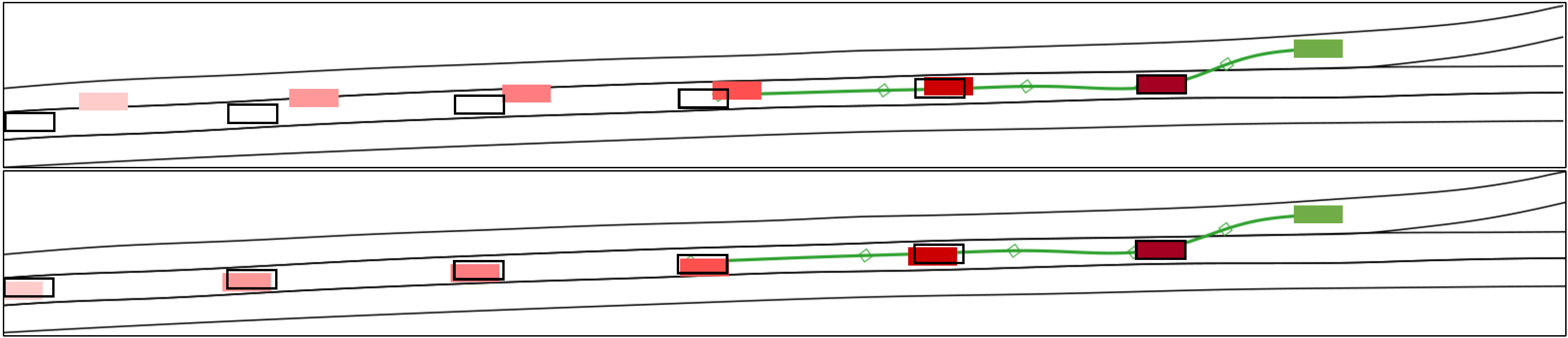}
\caption{Trajectory prediction and motion planning for an exemplar merging scenario from exiD dataset (top: Constant Velocity Prediction, bottom: Proposed POVL prediction). The green line shows the planned trajectory of the EV (green bounding box) for the simulation duration, with green diamonds representing one-second intervals for the next five seconds. The  red bounding box with black borders shows the current location of the vehicle on the main carriageway. The faded red bounding boxes and transparent bounding boxes with black borders represent predictions and ground truths in one-second intervals for the next five seconds, respectively.}
\label{fig: qual}
\end{figure*}

\begin{figure}[!t] 
\centering
\includegraphics[width=3.4in]{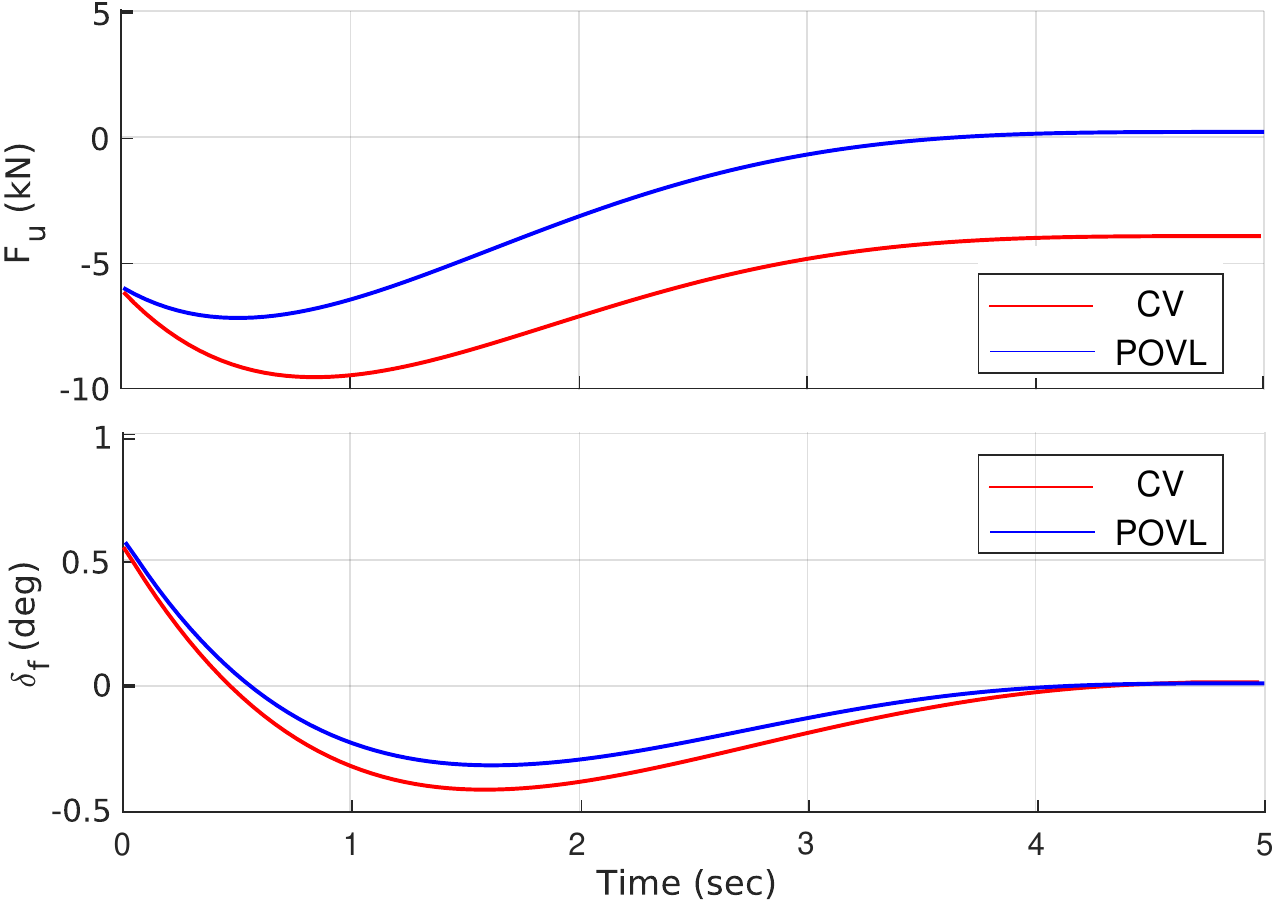}
\caption{Control effort signal (top: force, and bottom: steering angle) in motion planning using CV and POVL prediction methods for example scenario in Fig.~\ref{fig: qual}.}
\label{fig: qual_sig}
\end{figure}

\subsubsection{Comparison with baselines} Table~\ref{tab:comp_highd} compares the performance of the proposed model, some state-of-the-art highway trajectory prediction models, and Constant Velocity (CV) prediction evaluated on highD and exiD datasets. To the best of our knowledge, no trajectory prediction studies have been conducted using the exiD dataset. The selected state-of-the-art models include LSTM-based models such as MHA-LSTM~\cite{messaoud2020attention}, CNN-LSTM-based models such as CS-LSTM~\cite{deo2018convolutional},  and Transformer-based models such as~\cite{gao2023dual, mozaffari2020deep}. The results show that the proposed prediction model outperforms SOTA and CV models on both datasets in overall and specifically in longer prediction horizons. The CV predictor has a relatively low error and even outperforms one of the baseline learning-based models (e.g., CS-LSTM). We argue that this is because vehicles tend to keep their velocity in highway driving scenarios, therefore on average a CV prediction approach can achieve comparable results to learning-based methods on distance-based metrics such as RMSE.

\subsubsection{Performance in different observation lengths} Fig.~\ref{fig:ovl} shows the performance of the proposed prediction model evaluated on the exiD dataset for different observation lengths. The results show that the proposed method even with 0.4 seconds (i.e., 2-time steps) outperforms the CV model. Note that most existing learning-based prediction studies do not provide a prediction for observations less than 2-3 seconds.

\subsection{Prediction Impact on Planning} 
\subsubsection{Statistical Analysis} Table~\ref{tab:plan} shows the statistical analysis of the impact of trajectory prediction accuracy on motion planning and control metrics. We separately run motion planning algorithm with Ground-Truth (GT) futures of the vehicles as well as the CV and the proposed POVL models. The safety, comfort, and efficiency are reported for different traffic densities from the EV's perspective. To this end, the relevant metrics are calculated in seven sets, based on the max distance of the nearby vehicles to the EV's planned trajectory in each time step of the five-second simulation. The results show that using the proposed prediction model instead of CV can lead to a significant $9.15\%$ improvement in driving safety (i.e., $iTTC$) in total. However, the impact on the ride comfort (i.e., jerk) and fuel efficiency (i.e., Force)  is lower with values of $1.02\%$ and $4.80\%$, respectively. In denser traffic from EV's perspective (e.g., distance from other vehicles to EV less than 3 meters), the impact significantly increases to $19.18\%$, $19.78\%$, and $13.20~\%$ for safety, comfort, and efficiency, respectively. We argue that this is because although the average improvement in prediction accuracy of other vehicles is around 1 meter with POVL compared to CV, this improvement becomes relatively smaller in longer distances of the vehicle to the EV compared to shorter distances. Also, note that the magnitude of the potential field component of obstacle vehicles (i.e., $V_o$) in the motion planning cost function  is negligible for far vehicles.

\def\wb{0.5}
\renewcommand{\arraystretch}{1.7}
\begin{table} 
\centering
\caption{Comparison of different prediction approaches on comfort, efficiency, and safety of motion planning} \label{tab:plan}
\begin{tblr}{
  hline{1,15} = {-}{1pt},
  hline{3,7,11} = {-}{.5pt},
  hline{4,8,12} = {-}{.1pt},
  hline{2} = {3-9}{.2pt},
  cell{1}{1} = {r=2}{c},
  cell{1}{2} = {r=2}{c},
  cell{1}{3} = {c=7}{c},
  cell{3}{1} = {r=4}{c},
  cell{7}{1} = {r=4}{c},
  cell{11}{1} = {r=4}{c},
  column{1}={c}{.4cm},
  column{2}={c}{0.7cm},
  column{3-12}={c}{.5cm},
  row{4,8,12} = {.48cm},
  row{1-15} = {m},
}
\parbox[t]{2mm}{\rotatebox[origin=c]{90}{Metric}}          & Pred. Model & Distance from other vehicles to the EV (m) &       &       &      &      &      &      \\
                &                  & $<$3                  & $<$5     & $<$7     & $<$10   & $<$20   & $<$30   & $<$50   \\
\rotatebox[origin=c]{90}{\parbox{1.5cm}{\centering $iTTC \times 100$ ($\text{1/s}$)}} 
                                   & GT    & 11.71 & 17.13  & 16.59  & 16.74  & 16.83  & 17.64  & 18.54          \\
                                   & CV  & 14.96  & 21.12  & 19.96  & 18.60  & 18.63  & 19.51  & 20.42          \\
                                   & POVL     & 12.09  & 17.95  & 16.86  & 16.66  & 16.86  & 17.72  & 18.55          \\ 
                                   & Rel.\%*                        & 19.18 & 15.00 & 15.53 & 10.43  & 9.50  & 9.17  & 9.15          \\ 
\rotatebox[origin=c]{90}{\parbox{1.5cm}{\centering Jerk ($m/s^3$)}}          
                                   & GT                        & 4.32 & 4.41 & 4.49 & 4.76  & 4.57  & 4.64  & 4.80          \\ 
                                   & CV   & 5.61  & 5.23  & 5.05  & 4.93  & 4.65  & 4.69  & 4.87          \\
                                   & POVL                       & 4.50   & 4.50   & 4.55  & 4.77  & 4.57  & 4.63  & 4.82          \\
                                   & Rel.\%*                        & 19.78 & 13.95 & 9.90 & 3.24  & 1.72 & 1.27  & 1.02          \\ 
\rotatebox[origin=c]{90}{\parbox{1.5cm}{\centering Force  ($kN$)}}   
                                   & GT                                  & 10.95  & 8.85  & 8.88  & 8.82  & 8.77  & 8.89  & 8.97          \\ 
                                   & CV  & 12.65 & 10.25 & 9.61 & 8.82 & 8.84 & 8.90 & 9.13         \\
                                   & POVL                                & 10.98  & 8.87 & 9.01 & 8.81 & 8.80 & 8.92 & 8.99         \\
                                   & Rel.\%*                             & 13.20 & 13.46 & 6.24 & 0.11 &  0.45 &  -0.2 &  1.53         
\end{tblr}
\footnotesize{* The percentage of relative improvement of POVL compared to CV: (CV-POVL)/CV }

\end{table}

\subsubsection{Qualitative results}
Fig.~\ref{fig: qual} shows the trajectory prediction and motion planning results for an exemplary driving scenario in two cases of the CV and the POVL (proposed) prediction model. In this scenario, the EV is merging behind a vehicle. The CV model prediction contains a significant longitudinal error and a deviation to the right lane, which consequently results in sharper merging behaviour compared to a more accurate POVL prediction. In addition, the EV has to apply more negative force to avoid collision with the predicted vehicle in case of CV compared to POVL. Fig.~\ref{fig: qual_sig} depicts the control signals of motion planning for both prediction approaches.
\section{Conclusion}\label{sec:conc}
This paper presents a novel transformer-based prediction model with a variable-length observation window that enhances the accuracy of trajectory prediction in the highway driving scenario.
Through statistical analysis conducted on around 100 merges, it is illustrated that the improved trajectory prediction accuracy can significantly enhance the performance of a model predictive motion planner and improve safety, comfort, and fuel efficiency specifically in scenarios with closer distance of the EV to other vehicles. 
Future research endeavours will concentrate on implementing prediction and motion planning algorithms on sensory-equipped vehicles and addressing challenges related to real-world implementation, such as real-time performance and noisy sensor measurements.

\bibliographystyle{IEEEtran}

\bibliography{references}

\end{document}